\PassOptionsToPackage{dvipsnames}{xcolor}
\documentclass[lettersize,journal]{IEEEtran}
\usepackage{amsmath,amsfonts}
\usepackage{algorithmic}
\usepackage{algorithm}
\usepackage{array}
\usepackage[caption=false]{subfig}
\usepackage{textcomp}
\usepackage{stfloats}
\usepackage{url}
\usepackage{verbatim}
\usepackage{graphicx}
\usepackage{cite}
\usepackage[colorlinks=true,linkcolor=black,anchorcolor=black,citecolor=black,filecolor=black,menucolor=black,runcolor=black,urlcolor=black]{hyperref} 
\usepackage{booktabs}

\usepackage{ifthen}
\usepackage{tikz}
\usetikzlibrary{matrix,calc}
\usepackage{xcolor}

\begin{document}

\title{Wi-Fi based Human Fall and Activity Recognition using Transformer-based Encoder–Decoder and Graph Neural Networks}
\author{Younggeol Cho, Elisa Motta, Olivia Nocentini, Marta Lagomarsino, \\Andrea Merello, Marco Crepaldi, and Arash Ajoudani.

\thanks{The authors are with the HRI$^2$ Lab, Istituto Italiano di Tecnologia, Genoa, Italy, (Corresponding author: younggeol.cho@iit.it)}
}

\maketitle

\begin{abstract}
Human pose estimation and action recognition have received attention due to their critical roles in healthcare monitoring, rehabilitation, and assistive technologies. In this study, we proposed a novel architecture named \textit{Transformer-based Encoder–Decoder Network} (TED-Net) designed for estimating human skeleton poses from Wi-Fi Channel State Information (CSI). TED-Net integrates convolutional encoders with transformer-based attention mechanisms to capture spatiotemporal features from CSI signals. The estimated skeleton poses were used as input to a customized Directed Graph Neural Network (DGNN) for action recognition. We validated our model on two datasets: a publicly available multi-modal dataset for assessing general pose estimation, and a newly collected dataset focused on fall-related scenarios involving 20 participants. Experimental results demonstrated that TED-Net outperformed existing approaches in pose estimation, and that the DGNN achieves reliable action classification using CSI-based skeletons, with performance comparable to RGB-based systems. Notably, TED-Net maintains robust performance across both fall and non-fall cases. These findings highlight the potential of CSI-driven human skeleton estimation for effective action recognition, particularly in home environments such as elderly fall detection. In such settings, Wi-Fi signals are often readily available, offering a privacy-preserving alternative to vision-based methods, which may raise concerns about continuous camera monitoring.

\end{abstract}


\section{Introduction}

Estimating human skeleton pose is helpful in a wide range of fields, particularly in assistive technologies. Reconstructed skeleton keypoints not only facilitate precise motion tracking but also enable higher-level tasks such as skeleton-based action recognition. Among these tasks, fall detection plays a critical role in assisting vulnerable individuals, such as older adults or those with limited mobility, and alerting caregivers when immediate intervention is needed.

Traditionally, human skeleton reconstruction has been dominated by vision-based methods, many of which leverage deep learning. Well-known examples include YOLOpose \cite{yolo11_ultralytics}, OpenPose \cite{cao2019openpose}, and MediaPipe \cite{bazarevsky2020blazepose}, which are used to infer 2D joint positions from RGB images, often in real time. More recent solutions exploit multi-camera RGB imagery and machine learning algorithms to achieve more robust 3D human pose tracking \cite{realmove}. Beyond standard cameras, researchers have also explored LiDAR-based pose estimation to capture depth information and achieve more robust performance under challenging illumination conditions. For instance, LiveHPS \cite{ren2024livehps} can estimate 3D human pose and shape in large-scale environments, and LiDARCap \cite{zhang2024lidarcapv2} reconstructs 3D poses under human–object interaction.

Moreover, the reconstructed skeletal data have been proved particularly effective for fall detection. Several research groups have proposed approaches leveraging human skeleton pose information to detect falls, utilizing Long Short-Term Memory (LSTM) and Gated Recurrent Units (GRU) \cite{lin2020framework}, Attention-based LSTM \cite{bharathi2024real}, and Convolutional Neural Networks (CNN) \cite{xu2020fall}. 
Although vision-based systems offer accuracy and detailed visualization, they raise strong concerns regarding privacy, system complexity, and constraints such as line-of-sight requirements and sensitivity to environmental lighting, factors that can limit their feasibility in scenarios like fall detection in private or poorly lighted settings.

Recent advancements in wireless sensing have made WiFi Channel State Information (CSI) a better privacy-preserving alternative for human skeleton reconstruction. By capturing alterations in wireless signals caused by (human) movement, CSI-based solutions effectively overcome many of the limitations of traditional cameras, including lighting conditions and visual constraints. Moreover, since WiFi routers are already widely deployed in indoor environments, these systems require minimal additional hardware and can operate continuously without explicitly recording sensitive visual data \cite{kaur2024human}. Early approaches include CSI2Image \cite{kato2021csi2image}, which used Generative Adversarial Networks (GANs) to reconstruct images from CSI signals, and the teacher-student paradigm by Avola et al. \cite{avola2022human}, which generated both silhouette and skeleton sequences, translating CSI signals into videos. Additionally, transformer-based models have begun to appear in this domain, such as in the approach adopted by Zhou et al. \cite{zhou2023metafi++} for skeleton pose estimation.

CSI-based approaches are well-suited in controlled or semi-constrained environments, including bathrooms and changing rooms, where privacy is paramount. Such settings often offer stable conditions, like limited movement, fewer obstructions, and consistent furniture placement, which enhance the reliability of the Wi-Fi signal. Under these circumstances, continuous monitoring for events such as falls becomes feasible: a system can passively track user motion and alert caregivers when assistance is required. This context motivates our research goal of integrating skeleton pose estimation from CSI signals into a broader action recognition pipeline oriented toward fall detection.

This paper introduces a new \textit{Transformer-based Encoder–Decoder Network} (TED-Net) designed to estimate human skeleton poses from CSI signals. Those skeletons constitute the input to a customized Directed Graph Neural Network (DGNN) for skeleton-based action recognition, including fall detection. Unlike existing approaches, the DGNN model exclusively processes CSI-derived skeleton data, further enhancing privacy and reducing hardware requirements. To evaluate our proposed network, we conducted experiments on two distinct datasets: one from the literature for comparative analysis \cite{yang2023mm}, and another that we collected from 20 subjects to consider fall-included scenarios. The performance evaluation was focused on assessing both the accuracy of skeleton pose estimation and action recognition.

The primary contributions of this work are summarized as follows\footnote{The video of our work can be found at https://youtu.be/riaQshdg99Q}:
\begin{itemize}
\item A new architecture, TED-Net, designed for estimating human skeleton poses from Wi-Fi CSI signals, which outperforms state-of-the-art solutions;
\item A customized DGNN designed specifically for skeleton-based action recognition and fall detection;
\item An integrated framework that combines TED-Net and the customized DGNN to distinguish between voluntary, safe actions and falls that require alerting, using only WiFi signals; 
\item A preliminary artificial dataset addressing fall-related actions and comprehensively validating the proposed models. 
\end{itemize}

Exploiting these contributions for Wi-Fi-based fall and inactivity (e.g., due to loss of consciousness and stroke) recognition in home settings can expedite the detection of emergencies for the elderly, enabling faster intervention and potentially preventing the severe consequences of delayed assistance.

\begin{figure*}[t]
\centering
\centerline{\includegraphics[width=2\columnwidth]{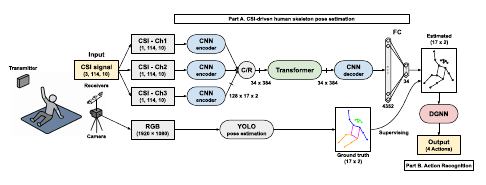}}
\caption{Overview of the CSI-driven Human Skeleton Pose Estimation and Action Recognition. The network consists of two main parts: (Part A) CSI-based human skeleton pose estimation (TED-Net) and (Part B) action recognition. CSI signals from three antennas pass through CNN encoders and a Transformer to estimate skeleton keypoints. Keypoints estimated by YOLO from RGB image supervise the training. Estimated poses feed into a DGNN for action recognition and fall detection. C: Concatenate, R: Reshape.}
\label{fig:overview}
\end{figure*}

\section{Methodology}
This section provides an overview of the network architecture, depicted in Fig. \ref{fig:overview}. The proposed method consists of two main components: the TED-Net (Network Part A) for human skeleton pose estimation from CSI signals, and a custom DGNN (Network Part B) for action recognition. The skeletal data produced by the TED-Net serves as input to the DGNN, which classifies actions into four categories: stand, walk, squat, and fall.

\begin{table}[!b]
\centering
\setlength{\tabcolsep}{15pt}
\renewcommand{\arraystretch}{1.2}
\caption{TED-Net structure and parameters.}
\begin{tabular}{l|c}
\hline\hline
Layer / Operation     & Output Shape \\
\hline\hline
Input                  & (1, 114, 10) \\
\hline
\textbf{CNN-Encoder 1}                  & (64, 58, 5) \\
(kernel=(4,3), stride=(2,2), padding=(2,1)) & \\
\textbf{CNN-Encoder 2}                  & (128, 31, 3) \\
(kernel=(2,3), stride=(2,2), padding=(2,1)) & \\
\textbf{CNN-Encoder 3}                  & (128, 17, 2) \\
(kernel=(2,3), stride=(2,2), padding=(2,1)) & \\
\hline
Concatenate (3 recievers) & (384, 17, 2) \\
Reshape                   & (34, 384) \\
\hline
\textbf{Transformer}               & (34, 384) \\
(2 layers, 8 heads)       & \\
\hline
\textbf{CNN-Decoder 1}             & (64, 68) \\
(kernel=3, stride=2, padding=1) & \\
\textbf{CNN-Decoder 2}             & (32, 136) \\
(kernel=3, stride=2, padding=1) & \\
\hline
Flatten                 & (4352) \\
Fully Connected Layer   & (34) \\
\hline
Reshape (Output)   & (17, 2) \\
\hline \hline
\end{tabular}
\label{tab:TED-Net}
\end{table}

\subsection{CSI-driven Human Skeleton Pose Estimation}
The proposed TED-Net estimates 17 bidimensional human pose keypoints directly from CSI signals. TED-Net combines the strengths of convolutional layers in extracting local spatio-temporal features with the global modeling capabilities of the Transformer self-attention mechanism. An overview of the architecture is shown in Table~\ref{tab:TED-Net}. The network receives CSI tensors, specifically one tensor per receiving antenna, of shape $(1,114,10)$. Each tensor is processed by a three-layer convolutional encoder that progressively reduces spatial dimensions while increasing channel depth, extracting the raw CSI signal into a compact feature representation. This hierarchical approach helps isolate relevant motion cues from noise. If multiple CSI tensors are available, for example, from different antennas, their encoded features are concatenated along the channel dimension to form a unified feature map.

Once concatenated, the feature map is reshaped into a sequence of length $34$ (i.e., $17 \times 2$) with an embedding dimension (e.g., $384$). This step effectively compresses the joint and coordinate dimensions into a single sequence axis, allowing the Transformer Encoder to capture global dependencies among spatially distant keypoints, as, for example, hand movements can strongly correlate with foot positions even though they are far apart on the body. The Transformer Encoder module (with two encoder layers and eight attention heads) applies self-attention across this entire sequence, enabling the network to learn rich, long-range interactions.

The Transformer Encoder outputs are then passed through a convolutional decoder. This decoder up-samples and refines the high-level features, reconstructing crucial spatial details of the human pose. A final fully connected layer, with a hyperbolic tangent activation function, maps the features into $34$ values, corresponding to $(x,y)$ coordinates for the 17 keypoints. The network reshapes this vector into $(17,2)$ to produce the final 2D skeleton.

TED-Net is trained using a Mean Squared Error (MSE) loss function on the estimated joint coordinates. We employ the Adam optimizer with a moderate batch size, $512$, for stable training dynamics. By design, the convolutional encoder focuses on localized spatiotemporal information, while the Transformer Encoder captures global, cross-joint interactions; the decoder then combines these signals into accurate 2D pose predictions.

\begin{table}[t]
\centering
\setlength{\tabcolsep}{8pt}
\renewcommand{\arraystretch}{1.3}
\caption{DGNN network structure and parameters.}
\begin{tabular}{l|c|c|r}
\hline\hline
\textbf{Layer} & \textbf{Input} & \textbf{Output} & \textbf{Params} \\
\hline\hline
data\_bn\_v (BN1d)        & (batch, 34)    & (batch, 34)     & 68 \\
data\_bn\_e (BN1d)        & (batch, 32)    & (batch, 32)     & 64 \\
\hline
\textbf{GraphTemporalConv 1} &               &                  & 3,184 \\
\quad DGNBlock (Linear)   & (6)            & (16)            & 832 \\
\quad TemporalConv        & (16)           & (16)            & 2,352 \\
\hline
\textbf{GraphTemporalConv 2} &               &                  & 17,824 \\
\quad DGNBlock (Linear)   & (48)           & (32)            & 3,808 \\
\quad TemporalConv        & (32)           & (32)            & 9,312 \\
\quad Residual Conv       & (16)           & (32)            & 4,704 \\
\hline
\textbf{GraphTemporalConv 3} &               &                  & 68,896 \\
\quad DGNBlock (Linear)   & (96)           & (64)            & 13,216 \\
\quad TemporalConv        & (64)           & (64)            & 37,056 \\
\quad Residual Conv       & (32)           & (64)            & 18,624 \\
\hline
Dropout1                  & (64)           & (64)            &  \\
Dropout2                  & (64)           & (64)            &  \\
ReLU                      & (64)           & (64)            &  \\
Fully Connected (FC)      & (128)          & (4), Classes             & 516 \\
\hline
\textbf{Total}            & —              & —               & \textbf{90,552} \\
\hline\hline
\end{tabular}
\label{tab:dgnn}
\end{table}

\subsection{Action Recognition and Fall Detection}
The DGNN is used as a graph-based approach to recognize human actions. The network directly employs skeletal data defined by keypoints and the predefined edges connecting them, which correspond to anatomical connections within the human body. DGNN leverages the skeleton's edge structure to propagate information effectively throughout the network, enhancing the interaction between directly connected keypoints. Specifically, it employs directed edges to preserve hierarchical connections among graph nodes. Additionally, the DGNN dynamically adjusts its adjacency matrix during training to emphasize connections among closely related keypoints.

In addition to spatial information from the graph structure, the DGNN also captures temporal dynamics through temporal convolution. Unlike the original DGNN approach described in \cite{shi2019skeleton}, which classifies an entire action sequence into a single action class, our approach requires precise temporal localization of events, such as falls. Therefore, we feed the model a temporal window containing the previous frames to classify the action in the current frame. As a result, our implementation enables frame-level recognition of actions. The details of the network structures are shown in Table \ref{tab:dgnn}.

The output classes (stand, walk, squat, and fall) were chosen to enable the network to learn the distinction between falling and non-falling scenarios effectively. In particular, \emph{squat} (or sitting) action can visually resemble an intermediate stage of a fall when viewed out of context. Including it as a separate class helps the network differentiate between voluntary, transitional, and safe postures and actual fall events, improving the overall robustness of fall detection.

\section{Experimental Campaign}

\subsection{Setup}
\begin{figure}[t]
\centering
\centerline{\includegraphics[width=0.9\columnwidth]{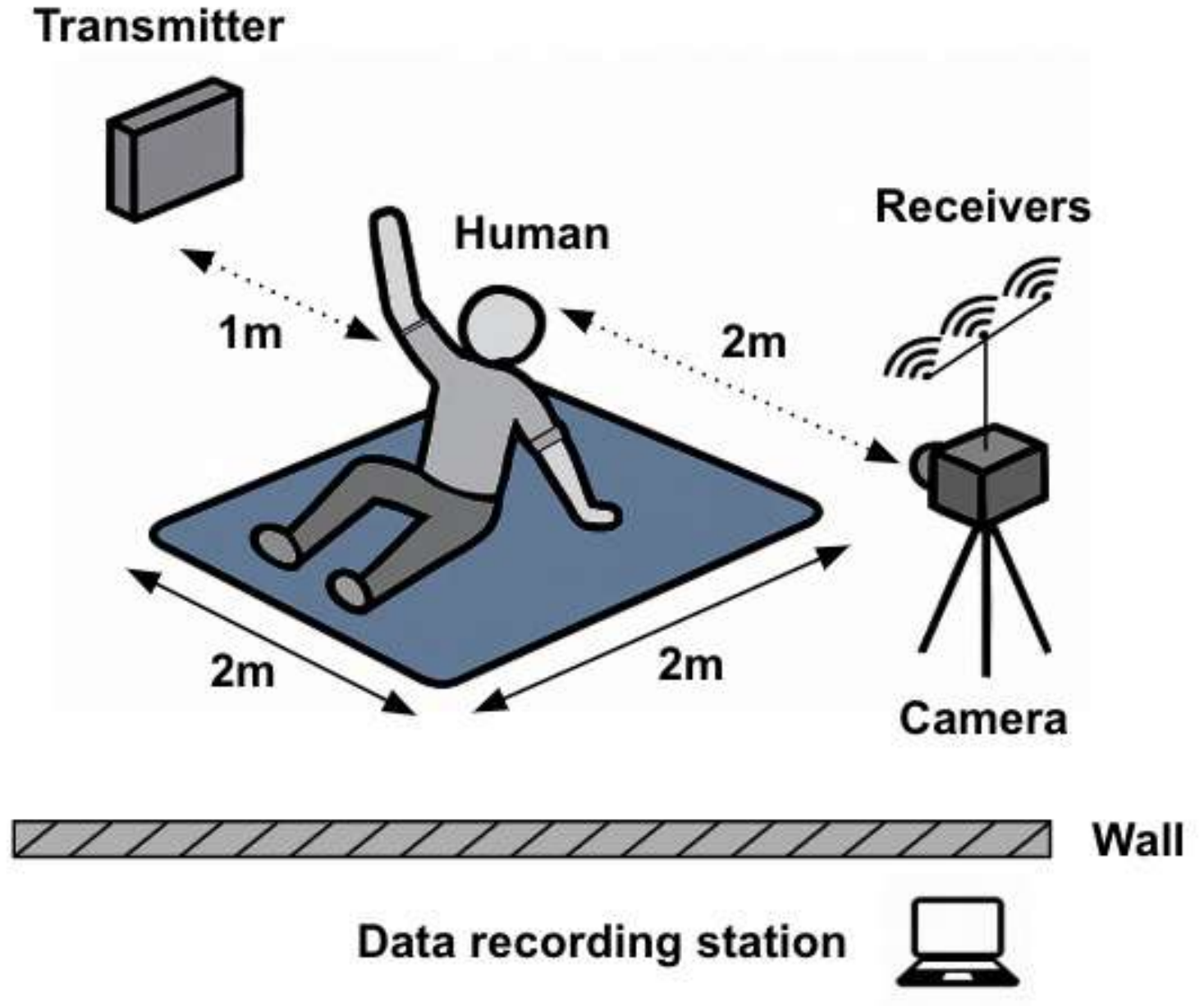}}
\caption{Experimental setup. The CSI signal was emitted from one transmitter and captured by three receivers. A camera was placed at the same location as the receivers. The subject performed actions on the mattress.}
\label{fig:setup}
\end{figure}

To simultaneously collect and synchronize CSI signals and corresponding images, we configured the experimental setup as illustrated in Fig. \ref{fig:setup}. A single transmitter was positioned on one side of the wall, while three antennas and one camera were installed on the opposite side. For safety, subjects performed all actions on a four-square-meter mattress, in two different locations: the center and left side. To capture changes in CSI signals, we utilized one transmitter and three receiver antennas, all custom-built using an ESP32S3 MCU board (Seeed Studio), running a custom-made firmware, and operating at a frequency of 2.4 GHz (same as common home WiFi frequency). The transmitter sends short WiFi packets at fixed rate, and the receivers extract the CSI data from those packets and send it over a USB/serial channel to a central PC.

The CSI data validity has been checked by connecting the transmitter and the receivers through coaxial cables, attenuators and RF-mixers, while injecting also a narrow-band interference signal. It has been verified that sweeping the WiFi 40MHz channel with the signal generator caused consistent results in the collected CSI data (i.e. the amplitude of proper OFDM subchannel is affected by the presence of the interfering signal). The transmitted WiFi packets contain a sequence number that helps to synchronize the data flow from the receivers; for this purpose we developed a custom PC program that is able to collect the data from all the receivers and gathers the measurements for each packet.

The system employed a single antenna configuration capable of data rates up to 150 Mbps. Each of the three receiving antennas collected CSI signals across 114 subcarriers at a sampling rate of 300 Hz. To facilitate signal processing, a buffer window of 10 samples was applied, resulting in input data with dimensions (10, 3, 114) and an effective acquisition rate of 30 Hz. CSI signals from all subcarriers were normalized by dividing by the maximum hardware specification value of 256, corresponding to the 8-bit resolution. Concurrently, a GoPro HERO11 camera captured wide-field RGB images at a resolution of 1920 × 1080 pixels, synchronized at a frame rate of 30 Hz to match the CSI data. YOLOv11Pose, a deep learning-based human pose estimation model, was applied to these RGB images to extract human skeleton keypoints. The YOLOv11Pose output consisted of 17 keypoints representing essential body joints and parts, which served as the ground truth for training our network. The keypoint coordinates were normalized by dividing by the image width and height. Additionally, when keypoints were occasionally missing or appeared significantly displaced compared to the previous frame, we applied interpolation, using keypoints from the previous frame and two frames prior, to fill in these gaps, ensuring smoother and more reliable keypoint trajectories. Data synchronization between the CSI signals and RGB images was managed using Robot Operating System 2 (ROS2).

\subsection{Datasets}
We validated our proposed network using two datasets: the MM-Fi dataset \cite{yang2023mm} and a custom dataset self-collected. Specifically, we assessed the TED-Net model's ability to estimate human skeletal poses from CSI signals using both datasets. However, for fall detection, we relied only on our custom dataset, as the external dataset did not include fall-related activities.
The MM-Fi dataset is a multi-modal human pose dataset that includes synchronized frames from RGB, infrared, depth, LiDAR, and CSI signals. It features 40 subjects performing 27 distinct daily and rehabilitation activities, such as walking, squatting, and lunging, across four different environments. To be consistent, for our experiments, we used a subset of this dataset consisting of 10 subjects performing all 27 activities in a single environment; each action sequence was split by duration into 80\% for training and 20\% for testing of the TED-Net.

Our custom dataset includes synchronized frames from RGB images and CSI signals recorded by three receivers. It was collected among 20 subjects (7 females and 13 males) with an average age of $\pm29$ years and an average height of $\pm172$ cm. The subjects performed five different actions: standing, walking, squatting, falling backward from a walking posture, and falling sideways from a squatting posture. To illustrate that the performance of reconstruction and action recognition is not location-dependent, and that falls can be differentiated from other actions such as walking and squatting, the subjects initiated walking and squatting from various positions (e.g., the left side of the mattress) before executing a fall.

For standing, walking, and squatting, data were collected continuously for 30 seconds each. We split these sequences by duration into 80\% for training the TED-Net and DGNN, and 20\% for testing the TED-Net. The CSI-based skeletons generated during testing, together with the corresponding RGB-skeletons, were used to evaluate the DGNN. 

For falling, each subject repeated the action five times, with each repetition comprising five seconds of walking or squatting followed by a five-second fall. The first four repetitions used RGB-skeletons for training both the TED-Net and the DGNN. The CSI signals from the fifth repetition were reserved as test data for the TED-Net. The resulting CSI-based skeletons, alongside the corresponding RGB-skeletons, were then used to evaluate the DGNN.

A detailed explanation of the experimental procedures and associated safety precautions was provided to all participants, and informed consent was obtained. All experiments were conducted in strict compliance with the ethical principles outlined in the Declaration of Helsinki. The research protocol was formally approved by the anonymous Ethics Committee.

\subsection{Analysis}
\subsubsection{CSI-driven Human Skeleton Pose estimation}
The performance of human pose estimation using CSI signals was evaluated using the widely adopted Percentage of Correct Keypoints (PCK) metric, which facilitates comparison with prior work on pose estimation from Wi-Fi signals \cite{zhou2023metafi++}. The PCK metric is defined as follows:
\begin{equation}
\label{PCK}
\centering
\text{PCK}_{\alpha} = \frac{1}{N}\sum_{i=1}^{N}\mathbb{I}(d_{i} \leq \alpha).
\end{equation}
$N$ represents the total number of frames being evaluated. The term $d_{i}$ is the Euclidean distance between the estimated keypoints and the ground-truth, and it was normalized by the diagonal length of the torso. $\alpha$ is a predefined fraction indicating the minimum required proportion of correctly detected joints within a skeleton for it to be considered correct. Finally, $\mathbb{I}$ is the indicator function, which equals 1 when the condition inside it is true and 0 otherwise.

\subsubsection{Action Recognition and Fall Detection}
Recognition accuracy was evaluated for five distinct human actions under two different evaluation scenarios: a multi-class classification scenario involving all five actions (represented as four distinct classes), and a binary classification scenario distinguishing between fall and non-fall cases. The results of the multi-class classification were also presented using a confusion matrix, providing a clear visualization of correctly identified actions and misclassifications.

\begin{figure*}[!b]
\centering
\centerline{\includegraphics[width=1.8\columnwidth]{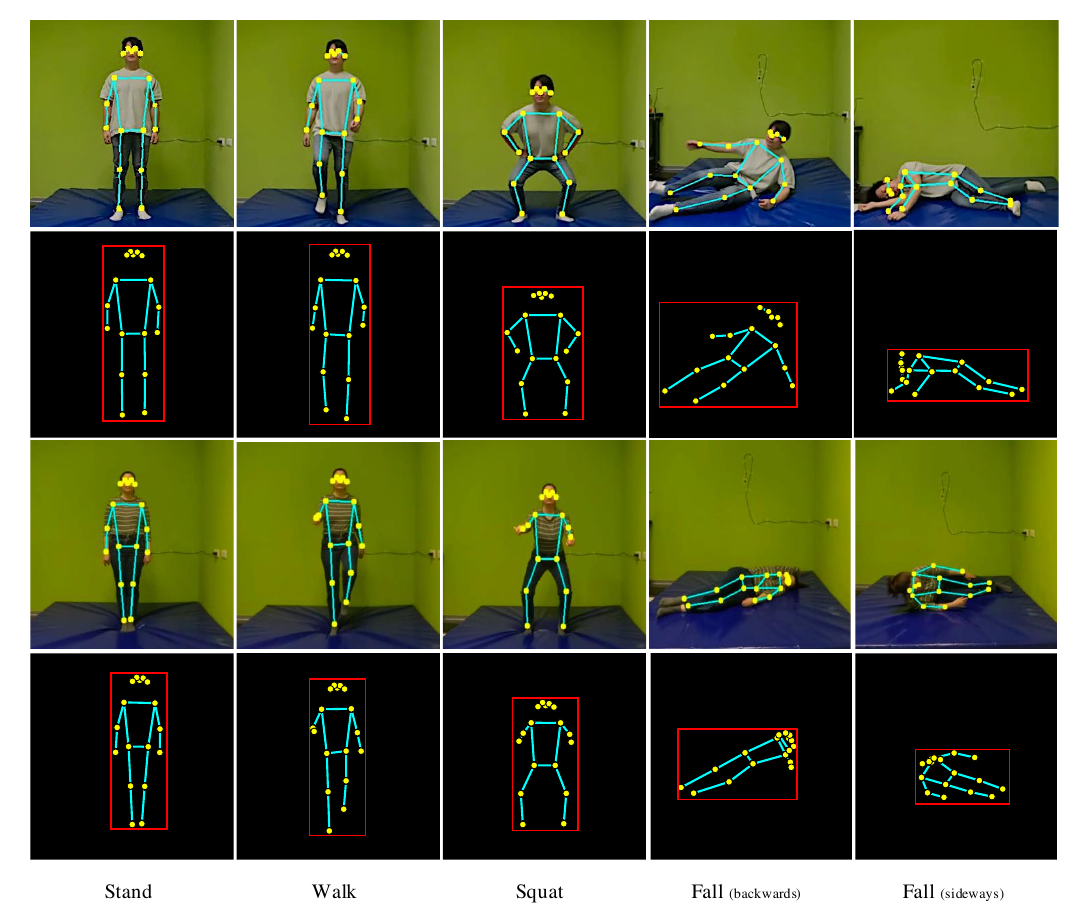}}
\caption{CSI-driven Human skeleton pose estimation result based on Ted-Net. This example illustrates two subjects performing the five actions, demonstrating accurate estimations for non-fall cases. Conversely, in fall scenarios, estimation errors are noticeable, particularly at the distal parts of the upper body.}
\label{fig:example}
\end{figure*}

\begin{table}[!t]
\centering
\setlength{\tabcolsep}{7pt}
\renewcommand{\arraystretch}{1.3}
\caption{External dataset PCK metrics}
\begin{tabular}{l|c|c|c|c|c}
\hline\hline
\textbf{} & {{PCK$_{10}$}} & {PCK$_{20}$} & {PCK$_{30}$} & {PCK$_{40}$} & {PCK$_{50}$} \\
\hline\hline
Nose        & 78.2 & 92.4 & 95.8 & 97.0 & 97.9 \\
Eye R       & 78.9 & 92.6 & 95.9 & 97.0 & 97.9 \\
Eye L       & 78.9 & 92.6 & 95.8 & 97.1 & 97.9 \\
Ear R       & 82.7 & 94.1 & 96.6 & 97.7 & 98.4 \\
Ear L       & 82.8 & 94.1 & 96.6 & 97.7 & 98.3 \\
Shoulder R  & 85.0 & 95.4 & 98.2 & 99.1 & 99.6 \\
Shoulder L  & 84.5 & 95.2 & 98.0 & 99.0 & 99.6 \\
Elbow R     & 54.4 & 75.3 & 84.4 & 89.7 & 93.4 \\
Elbow L     & 53.7 & 75.6 & 85.5 & 90.5 & 93.6 \\
Hand R      & 37.3 & 57.7 & 69.0 & 76.4 & 81.5 \\
Hand L      & 37.9 & 58.4 & 69.8 & 77.4 & 82.3 \\
Pelvis R    & 99.3 & 99.9 & 100 & 100 & 100 \\
Pelvis L    & 99.3 & 99.9 & 100 & 100 & 100 \\
Knee R      & 81.6 & 92.7 & 96.3 & 98.0 & 99.0 \\
Knee L      & 81.9 & 93.1 & 96.8 & 98.3 & 99.3 \\
Foot R      & 68.9 & 84.5 & 92.0 & 95.1 & 96.8 \\
Foot L      & 68.7 & 84.9 & 92.1 & 95.5 & 97.2 \\
\hline
\textbf{AVG} & \textbf{73.8} & \textbf{87.0} & \textbf{91.9} & \textbf{94.5} & \textbf{96.0} \\
\hline
Literature \cite{zhou2023metafi++} & 63.5 & 86.7	& 93.1 & 95.9 & 97.3 \\
Literature \cite{yang2022metafi} & 64.2 & 83.2	& 90.0 & 93.3 & 95.2 \\

\hline\hline
\end{tabular}
\label{tab:external_PCK}
\end{table}

\section{Results}
\subsection{CSI-driven Human Skeleton Pose estimation}

The PCK scores were calculated for each keypoint across all actions performed by all subjects. Table \ref{tab:external_PCK} summarizes the PCK values obtained when applying TED-Net to the MM-Fi dataset \cite{yang2023mm}. The PCK was evaluated from PCK$_{10}$ to PCK$_{50}$, and the average values were compared against existing literature. TED-Net achieved an average PCK$_{10}$ score of 73.8\%, surpassing the results reported in the previous study \cite{zhou2023metafi++}, which achieved a score of 63.5\%. From PCK$_{20}$ to PCK$_{50}$, the PCK of both TED-Net and Literature \cite{zhou2023metafi++} showed the similar values that was over 90\% from PCK$_{30}$. Moreover, compared to Yang et al. study \cite{yang2022metafi}, TED-Net achieved superior results across all PCK metrics. Notably, keypoints located closer to the body center, such as the shoulders and pelvis, were estimated more accurately, as indicated by higher singular PCK$_{10}$ values. Conversely, accuracy decreased for keypoints farther from the body center, such as the hands and feet. Specifically, the highest accuracy was observed for the pelvis (99.3\%) and shoulders (approximately 85\%), while accuracy was notably lower for the hands (approximately 37\%).

Table \ref{tab:PCK_fall} summarizes the PCK scores for both non-fall ($X$ columns) and fall ($O$ columns) events when applying the TED-Net on our custom dataset. Specifically, $X$ denotes the network’s performance on standing, walking, and squatting, while $O$ reflects its performance on actual fall sequences. 
Notably, the PCK scores for non-fall actions in our custom dataset were consistently higher compared to those from the external dataset (Table \ref{tab:external_PCK}); for example, the average PCK$_{10}$ was 83.4\%, compared to 73.8\% in the external dataset, highlighting the network's robust performance across different datasets and scenarios. The results point out lower PCK values for non-fall actions across all PCK thresholds compared to fall events. Despite this lower overall accuracy, the reconstructed keypoints always generated a skeleton shape compatible with the performed action. Fig. \ref{fig:example} gives a sample of this, comparing the ground truth skeleton poses, derived from RGB images, with CSI-based skeleton estimations, generated by TED-Net, for two subjects. This example demonstrates that TED-Net is not subject-dependent; its performance is independent of individual subject characteristics. Despite the noticeable differences in body size and shape between the two subjects shown in the figure, TED-Net achieves accurate reconstruction results. Results suggest that even the relatively lower accuracy during fall actions can still serve effectively as indicative cues for recognizing falls.

\begin{table}[!t]
\centering
\setlength{\tabcolsep}{3pt}
\renewcommand{\arraystretch}{1.36}
\caption{Our dataset fall events PCK metrics}
\begin{tabular}{l|cc|cc|cc|cc|cc}
\hline\hline
& \multicolumn{2}{c|}{{PCK$_{10}$}} & \multicolumn{2}{c|}{{PCK$_{20}$}} & \multicolumn{2}{c|}{{PCK$_{30}$}} & \multicolumn{2}{c|}{{PCK$_{40}$}} & \multicolumn{2}{c}{{PCK$_{50}$}} \\
\hline\hline
Fall    & X & O  & X & O  & X & O  & X & O  & X & O  \\
\hline
Nose       & 81.6 & 30.2 & 92.1 & 49.6 & 95.8 & 57.9 & 98.3 & 63.5 & 99.4 & 71.7 \\
Eye R      & 81.7 & 31.5 & 92.3 & 51.0 & 96.1 & 59.4 & 98.4 & 64.8 & 99.4 & 73.0 \\
Eye L      & 82.0 & 35.6 & 92.2 & 57.7 & 95.9 & 66.2 & 98.2 & 71.9 & 99.4 & 78.8 \\
Ear R      & 83.2 & 33.4 & 92.8 & 51.6 & 96.6 & 58.6 & 98.8 & 65.0 & 99.5 & 74.0 \\
Ear L      & 83.0 & 39.4 & 92.4 & 64.1 & 96.3 & 73.6 & 98.6 & 79.9 & 99.4 & 85.0 \\
Shoulder R & 85.1 & 26.8 & 93.7 & 43.5 & 97.2 & 51.2 & 99.4 & 60.8 & 99.9 & 73.9 \\
Shoulder L & 85.1 & 25.9 & 93.7 & 44.5 & 97.3 & 52.1 & 99.4 & 63.2 & 99.9 & 74.7 \\
Elbow R    & 83.1 & 27.0 & 93.7 & 47.5 & 97.6 & 58.6 & 99.5 & 67.5 & 99.9 & 75.7 \\
Elbow L    & 82.6 & 28.1 & 94.4 & 50.7 & 97.9 & 62.9 & 99.8 & 74.6 & 99.9 & 82.4 \\
Hand R     & 70.3 & 19.8 & 86.7 & 43.9 & 93.7 & 55.6 & 97.2 & 64.9 & 98.6 & 73.0 \\
Hand L     & 70.6 & 20.9 & 87.7 & 47.5 & 95.1 & 62.5 & 97.8 & 75.8 & 98.9 & 84.2 \\
Pelvis R   & 90.1 & 28.0 & 96.8 & 50.6 & 99.5 & 74.7 & 100.0 & 89.3 & 100.0 & 95.0 \\
Pelvis L   & 89.6 & 28.2 & 96.7 & 54.9 & 99.5 & 75.9 & 100.0 & 91.4 & 100.0 & 95.4 \\
Knee R     & 91.1 & 27.2 & 97.9 & 58.6 & 99.0 & 75.6 & 99.7 & 84.8 & 100.0 & 91.2 \\
Knee L     & 89.4 & 28.3 & 97.6 & 65.5 & 98.9 & 80.9 & 99.4 & 89.5 & 99.6 & 91.9 \\
Foot R     & 84.8 & 26.7 & 94.0 & 63.3 & 97.3 & 75.9 & 98.6 & 83.6 & 99.3 & 87.3 \\
Foot L     & 84.3 & 28.3 & 93.2 & 66.0 & 96.6 & 78.6 & 98.3 & 85.4 & 99.1 & 89.6 \\ \hline
\textbf{AVG} & \textbf{83.4} & \textbf{28.5} & \textbf{93.4} & \textbf{53.6} & \textbf{97.1} & \textbf{65.9} & \textbf{98.9} & \textbf{75.1} & \textbf{99.5} & \textbf{82.2} \\
\hline\hline
\end{tabular}
\label{tab:PCK_fall}
\end{table}

\subsection{Action Recognition and Fall Detection}

Fig. \ref{fig:confusion} shows the confusion matrices comparing classification over RGB-skeletons, Fig. \ref{fig:a}, and CSI-skeletons, Fig. \ref{fig:b}. Table \ref{tab:accuracy} reports the accuracy values for both RGB and CSI obtained skeletons. Both RGB- and CSI-based methods showed similar recognition accuracy across the evaluated classes. Specifically, the CSI-based method achieved the average accuracy of 94.8\% (Stand: 98.0\%, Walk: 91.0\%, Squat: 95.4\%), while the RGB-based skeleton showed an average accuracy of 98.3\% for non-fall actions (Stand: 100\%, Walk: 98.3\%, Squat: 96.5\%). 

\begin{table}[!t]
\centering
\setlength{\tabcolsep}{10pt}
\renewcommand{\arraystretch}{1.4}
\caption{Recognition accuracy based on RGB and CSI skeletons.}
\label{tab:accuracy}
\begin{tabular}{c|c|c|c|c}
\hline\hline
              & STAND    & WALK    & SQUAT    & FALL \\
\hline
RGB-skeletons & 100      & 98.3    & 96.5     & 94.3 \\
\hline
CSI-skeletons & 98.0       & 91.0    & 95.4     & 90.5 \\
\hline\hline
\end{tabular}
\end{table}

\begin{figure}[!t]
\centering
\subfloat[\label{fig:a}]{
    \begin{tikzpicture}
    \begin{scope}[scale=0.8]

    \def\myConfMat{{   
        {240, 0,   0,   0},
        {0,   472, 0,   8},
        {1,   0,   463, 16},
        {0,   2,   23, 415},
    }}
    \def\classNames{{"\scriptsize{0}","\scriptsize{1}", "\scriptsize{2}", "\scriptsize{3}"}}
    \def\numClasses{4}

    \foreach \y in {1,...,\numClasses} {
      \node [anchor=east] at (-0.5,-\y) {\pgfmathparse{\classNames[\y-1]}\pgfmathresult};

      \foreach \x in {1,...,\numClasses} {
          \def\totSamples{0}
          \foreach \ll in {1,...,\numClasses} {
            \pgfmathparse{\myConfMat[\y-1][\ll-1]}   
            \xdef\totSamples{\totSamples+\pgfmathresult}
            }
          \pgfmathparse{\totSamples} \xdef\totSamples{\pgfmathresult}

          \begin{scope}[shift={(-1+\x,-\y)}]
              \def\mVal{\myConfMat[\y-1][\x-1]}
              \pgfmathtruncatemacro{\r}{\mVal}
              \pgfmathtruncatemacro{\p}{round(\r/\totSamples*100)}

              \pgfmathtruncatemacro{\pScaled}{ifthenelse(\p*2 > 100, 100, \p*2)}
            
                \ifthenelse{\pScaled<50}{\def\txtcol{black}}{\def\txtcol{white}}
                
              \node[
                  draw,
                  text=\txtcol,
                  align=center,
                  fill=OliveGreen!\pScaled,
                  minimum size=8mm,
                  font=\scriptsize,
                  inner sep=0
              ] (C) {\r\\\p\%};

              \ifthenelse{\y=\numClasses}{
                  \node [] at ($(C)-(0,0.8)$)
                    {\pgfmathparse{\classNames[\x-1]}\pgfmathresult};}{}
          \end{scope}
      }
    }
    \coordinate (yaxis) at (-1.2,0.5-\numClasses/2);
    \coordinate (xaxis) at (-0.3+\numClasses/2, -\numClasses-1.3);
    \node [rotate=90,anchor=east] at (yaxis) {\small{Target Classes}};
    \node [] at (xaxis) {\small{Predicted Classes}};
    
    \end{scope}
    \end{tikzpicture}
    \label{fig:conf1}
}
\hfil
\subfloat[\label{fig:b}]{
    \begin{tikzpicture}
    \begin{scope}[scale=0.8]

    \def\myConfMat{{
       {236, 3,   1,  0},
       {11,   437, 30,   2},
       {0,   17,   458, 5},
       {0,   11,   30,  399},
    }}
    \def\classNames{{"\scriptsize{0}","\scriptsize{1}", "\scriptsize{2}", "\scriptsize{3}"}}
    \def\numClasses{4}

    \foreach \y in {1,...,\numClasses} {
        \node [anchor=east] at (-0.5,-\y) {\pgfmathparse{\classNames[\y-1]}\pgfmathresult};

        \foreach \x in {1,...,\numClasses} {
            \def\totSamples{0}
            \foreach \ll in {1,...,\numClasses} {
                \pgfmathparse{\myConfMat[\y-1][\ll-1]}   
                \xdef\totSamples{\totSamples+\pgfmathresult}
            }
            \pgfmathparse{\totSamples} \xdef\totSamples{\pgfmathresult}

            \begin{scope}[shift={(-1+\x,-\y)}]
                \def\mVal{\myConfMat[\y-1][\x-1]}
                \pgfmathtruncatemacro{\r}{\mVal}
                \pgfmathtruncatemacro{\p}{round(\r/\totSamples*100)}

                \pgfmathtruncatemacro{\pScaled}{ifthenelse(\p*2 > 100, 100, \p*2)}

                \ifthenelse{\pScaled<50}{\def\txtcol{black}}{\def\txtcol{white}}
                \node[
                    draw,
                    text=\txtcol,
                    align=center,
                    fill=RoyalPurple!\pScaled,
                    minimum size=8mm,
                    font=\scriptsize,
                    inner sep=0
                ] (C) {\r\\\p\%};

                \ifthenelse{\y=\numClasses}{
                    \node [] at ($(C)-(0,0.8)$)
                      {\pgfmathparse{\classNames[\x-1]}\pgfmathresult};}{}
            \end{scope}
        }
    }

    \coordinate (yaxis) at (-1.2,0.5-\numClasses/2);
    \coordinate (xaxis) at (-0.3+\numClasses/2, -\numClasses-1.3);
    \node [rotate=90,anchor=east] at (yaxis) {\small{Target Classes}};
    \node [] at (xaxis) {\small{Predicted Classes}};
    \end{scope}
    \end{tikzpicture}
    \label{fig:conf2}
}
\caption{Confusion matrices showing recognition results based on (a) RGB image-based and (b) CSI-driven skeletons. 0:stand, 1: walk, 2:squat, 3:fall.}
\label{fig:confusion}
\end{figure}

For fall events, the CSI-based skeleton showed a comparable accuracy of 90.5\% to the RGB-based skeleton, which achieved an accuracy of 94.3\%. Additionally, when analyzing fall-specific misclassifications, the CSI-based skeleton resulted in 2.5\% false negatives labeled as walking and 6.8\% as squatting. Conversely, the false positive rates for the RGB-based skeleton results were 1.8\% for the label walk and 3.6\% for squat, while the CSI-based skeleton results showed lower values, 0.5\% for walk and 1.2\% for squat.

\section{Discussion}

As presented in Table \ref{tab:external_PCK}, TED-Net achieves better estimation performances compared to the models proposed by Zhou et al.\cite{zhou2023metafi++} and Yang et al.\cite{yang2022metafi}, evaluated on the external dataset \cite{yang2023mm} comprising exclusively non-fall events. TED-Net also demonstrated high accuracy on our custom dataset, which includes both fall and non-fall cases. These results suggest that, when trained on a comprehensive dataset covering a wide range of scenarios, TED-Net can perform reliably under semi-constrained conditions.

Analyzing fall-specific events portrayed by our custom dataset, the PCK scores at low $\alpha$ thresholds are significantly worse when only falling events are considered, as shown in Table \ref{tab:PCK_fall}. This is mostly due to the inherent characteristics of CSI signals, as they offer lower spatial resolution than camera-based data \cite{guo2019signal} and are susceptible to background noise and multipath reflections \cite{wang2016wifall,khan2020differential, peng2024wi}. In addition, rapid and high-amplitude movements, such as falling, further complicate the estimation. Once a person reaches the ground and ceases to move, the network receives negligible signal variation, making it difficult to refine the final lying-down pose \cite{wu2015non}. As a result, the reconstructed skeleton may remain in an intermediate position rather than reflecting a fully supine or prone orientation, leading to partial or incomplete fall depiction in the estimated pose.

However, it is important to note that this limitation in fine-grained pose accuracy does not significantly impact downstream action recognition. CSI-driven skeletons proved sufficiently robust to replace RGB-based skeletons for fall detection applications. The DGNN achieved comparable classification performance when using CSI-based skeletons versus RGB-based ones, indicating no significant differences, as shown in the confusion matrix in Fig. \ref{fig:confusion}. This proves that CSI-based skeletons are effective in capturing the overall dynamics of large-scale actions, even if some finer joint details are missing. 

\begin{figure}[!h]
\centering
\centerline{\includegraphics[width=1.0\columnwidth]{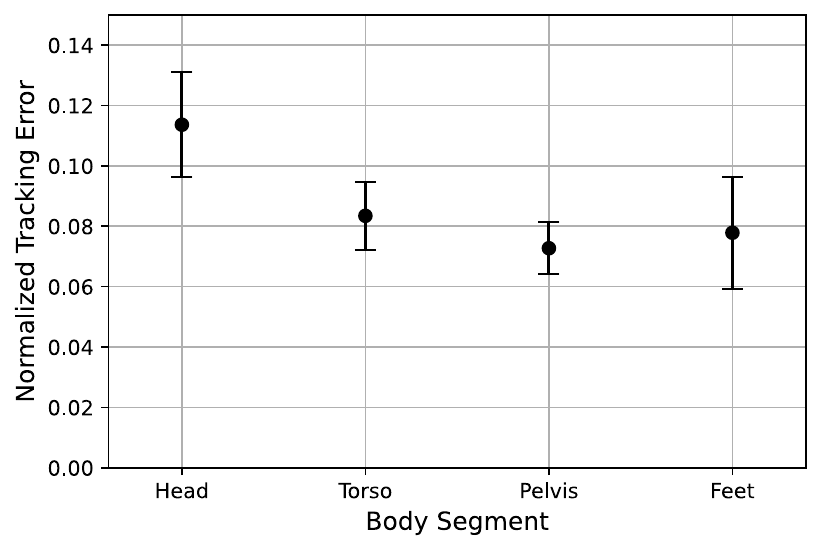}}
\caption{Normalized Tracking Error Across Body Segments. The position of each body segment was calculated as the average of the keypoints that comprise that segment.}
\label{fig:tracking}
\end{figure}

In addition, TED-Net also has the capability to track human positions. As illustrated in Fig. \ref{fig:tracking}, TED-Net can track the position of the human torso or pelvis with less than 10\% error, normalized against the diagonal length of the torso. This capability indicates that, in addition to recognizing human actions, TED-Net can monitor human positions for global activity analysis. As demonstrated in Table \ref{tab:external_PCK} and \ref{tab:PCK_fall}, tracking errors at the extremities of the body, such as the hands and feet, are slightly higher than those at the body's center (torso and pelvis). Nevertheless, as we can see from the result of human action recognition (see Fig. \ref{fig:confusion}), CSI-based skeleton tracking effectively captures overall body dynamics, enabling accurate recognition of human actions.

Nonetheless, open challenges need to be addressed in future work. The resolution of the CSI signal is constrained by hardware factors, as the number of emitters, antennas, and subcarriers. Moreover, estimation accuracy is influenced by environmental variability; changes such as room layout adjustments, furniture rearrangement, or other people moving around can introduce signal distortion and degrade pose estimation quality.  These limitations make it difficult to accurately track small body movements, such as hand gestures and foot positions. While our system showed robustness in action classification, improving fine-grained pose accuracy remains a key research direction. Therefore, maintaining consistent environmental conditions and carefully designing the signal acquisition system are key factors in achieving high-quality CSI signals and accurate human skeleton pose estimation.

\section{Conclusion}
In this paper, we introduced an end-to-end framework for vision-free human skeleton pose estimation and action recognition using Wi-Fi Channel State Information (CSI), with two core components: the Transformer-based Encoder–Decoder Network (TED-Net) and a custom Directed Graph Neural Network (DGNN). The performance was validated on both a dataset from the literature and a newly collected, fall-specific dataset. The result showed that TED-Net consistently outperforms existing CSI-based pose estimation models. Notably, the DGNN was able to classify human actions with high accuracy using only CSI-derived skeletons, comparable to the performance of RGB-based skeletons. In future work, TED-Net can be extended to explore 3D pose reconstruction from CSI and simultaneously support multiple individuals. Overall, this study highlights the potential of CSI-driven sensing for human motion tracking and activity recognition, suggesting possibilities for vision-free, privacy-preserving systems and creating new opportunities in the field.

\bibliographystyle{IEEEtran}
\bibliography{bibliography.bib}
\end{document}